\documentclass[10pt,twocolumn,letterpaper]{article}

\usepackage{cvpr}              % To produce the CAMERA-READY version
\usepackage{adjustbox}
\usepackage{multirow}
\usepackage{tabularx}

\usepackage[accsupp]{axessibility} % Improves PDF readability for those with disabilities.

\definecolor{cvprblue}{rgb}{0.21,0.49,0.74}
\usepackage[pagebackref,breaklinks,colorlinks,citecolor=cvprblue]{hyperref}

\def\paperID{*****} % *** Enter the Paper ID here
\def\confName{CVPR}
\def\confYear{2025}

\title{NTIRE 2025 Challenge on Image Super-Resolution ($\times$4): Methods and Results}
\author{
Zheng Chen$^{\dagger}$ \and
Kai Liu$^{\dagger}$ \and
Jue Gong$^{\dagger}$ \and
Jingkai Wang$^{\dagger}$ \and
Lei Sun$^{\dagger}$ \and
Zongwei Wu$^{\dagger}$ \and
Radu Timofte$^{\dagger}$ \and
Yulun Zhang$^{\dagger*}$ \and
Xiangyu Kong \and
Xiaoxuan Yu \and
Hyunhee Park \and
Suejin Han \and
Hakjae Jeon \and
Dafeng Zhang \and
Hyung-Ju Chun \and
Donghun Ryou \and
Inju Ha \and
Bohyung Han \and
Lu Zhao \and
Yuyi Zhang \and
Pengyu Yan \and
Jiawei Hu \and
Pengwei Liu \and
Fengjun Guo \and
Hongyuan Yu \and
Pufan Xu \and
Zhijuan Huang \and
Shuyuan Cui \and
Peng Guo \and
Jiahui Liu \and
Dongkai Zhang \and
Heng Zhang \and
Huiyuan Fu \and
Huadong Ma \and
Yanhui Guo \and
Sisi Tian \and
Xin Liu \and
Jinwen Liang \and
Jie Liu \and
Jie Tang \and
Gangshan Wu \and
Zeyu Xiao \and
Zhuoyuan Li \and
Yinxiang Zhang \and
Wenxuan Cai \and
Vijayalaxmi Ashok Aralikatti \and
Nikhil Akalwadi \and
G Gyaneshwar Rao \and
Chaitra Desai \and
Ramesh Ashok Tabib \and
Uma Mudenagudi \and
Marcos V. Conde \and
Alejandro Merino \and
Bruno Longarela \and
Javier Abad \and
Weijun Yuan \and
Zhan Li \and
Zhanglu Chen \and
Boyang Yao \and
Aagam Jain \and
Milan Kumar Singh \and
Ankit Kumar \and
Shubh Kawa \and
Divyavardhan Singh \and
Anjali Sarvaiya \and
Kishor Upla \and
Raghavendra Ramachandra \and
Chia-Ming Lee \and
Yu-Fan Lin \and
Chih-Chung Hsu \and
Risheek V Hiremath \and
Yashaswini Palani \and
Yuxuan Jiang \and
Qiang Zhu \and
Siyue Teng \and
Fan Zhang \and
Shuyuan Zhu \and
Bing Zeng \and
David Bull \and
Jingwei Liao \and
Yuqing Yang \and
Wenda Shao \and
Junyi Zhao \and
Qisheng Xu \and
Kele Xu \and
Sunder Ali Khowaja \and
Ik Hyun Lee \and
Snehal Singh Tomar \and
Rajarshi Ray \and
Klaus Mueller \and
Sachin Chaudhary \and
Surya Vashisth \and
Akshay Dudhane \and
Praful Hambarde \and
Satya Naryan Tazi \and
Prashant Patil \and
Santosh Kumar Vipparthi \and
Subrahmanyam Murala \and 
Bilel Benjdira \and
Anas M. Ali \and
Wadii Boulila \and
Zahra Moammeri \and
Ahmad Mahmoudi-Aznaveh \and
Ali Karbasi \and
Hossein Motamednia \and
Liangyan Li \and
Guanhua Zhao \and 
Kevin Le \and 
Yimo Ning \and 
Haoxuan Huang \and 
Jun Chen \and
}

\begin{document}

\maketitle

\let\thefootnote\relax\footnotetext{$^{\dagger}$Zheng Chen, Kai Liu, Jue Gong, Jingkai Wang, Lei Sun, Zongwei Wu, Radu Timofte, and Yulun Zhang are the challenge organizers, while the other authors participated in the challenge. $^{*}$Corresponding author: Yulun Zhang. Section~B in the supplementary materials contains the authors' teams and affiliations. NTIRE 2025 webpage: \url{https://cvlai.net/ntire/2025}. Code: \url{https://github.com/zhengchen1999/NTIRE2025_ImageSR_x4}.}

\vspace{-10.mm}
\begin{abstract}
This paper presents the NTIRE 2025 image super-resolution ($\times$4) challenge, one of the associated competitions of the 10th NTIRE Workshop at CVPR 2025. The challenge aims to recover high-resolution (HR) images from low-resolution (LR) counterparts generated through bicubic downsampling with a $\times$4 scaling factor. The objective is to develop effective network designs or solutions that achieve state-of-the-art SR performance. To reflect the dual objectives of image SR research, the challenge includes two sub-tracks: \textbf{(1) a restoration track}, emphasizes pixel-wise accuracy and ranks submissions based on PSNR; \textbf{(2) a perceptual track}, focuses on visual realism and ranks results by a perceptual score. A total of 286 participants registered for the competition, with 25 teams submitting valid entries. This report summarizes the challenge design, datasets, evaluation protocol, the main results, and methods of each team. The challenge serves as a benchmark to advance the state of the art and foster progress in image SR.
\end{abstract}

%% narrow the gap between equations and sentences
\setlength{\abovedisplayskip}{1pt}
\setlength{\belowdisplayskip}{1pt}

\vspace{-6mm}
\section{Introduction}
\vspace{-2mm}
Single image super-resolution (SR) refers to the process of reconstructing a high-resolution (HR) image from its low-resolution (LR) counterpart, which has undergone an information-losing degradation process. As a fundamental problem in computer vision, SR underpins numerous practical applications, including video surveillance, digital pathology, and satellite imagery. Among the various SR formulations, the classical image SR task~\cite{timofte2017ntire,zhang2023ntire} is one of the most commonly adopted benchmarks.

In this setting, the LR image is typically generated through a predefined downsampling operation, most often bicubic interpolation. Such degradation removes significant high-frequency content, rendering the task of detail recovery an inherently ill-posed inverse problem. The goal is to recover the missing high-frequency components using learned priors from external data. Owing to its well-defined setting, the classical SR task is a standard benchmark, enabling fair comparison across methods. Moreover, models trained under this framework often generalize well to more complex degradations through transfer learning.

Early image super-resolution (SR) methods rely on interpolation or reconstruction-based algorithms~\cite{zibetti2007robust,sun2010context,zhang2012single}. However, these approaches often produce over-smoothed results that lack fine textures and realistic details. With the advent of deep learning, convolutional neural networks (CNNs) have become the dominant paradigm in SR~\cite{dong2014learning,kim2016accurate,zhang2018image}. Beginning with the pioneering work of SRCNN~\cite{dong2014learning}, later models incorporate deeper architectures~\cite{wang2015deep,zhang2018image}, residual learning~\cite{zhang2018residual}, and attention 
mechanisms~\cite{dai2019second} to improve reconstruction accuracy. Meanwhile, Transformer-based architectures further advance SR by modeling long-range dependencies through self-attention~\cite{vaswani2017attention,liu2021swin,chen2023dual}. Besides, recent state-space models (\ie, Mamba) offer improved scalability and efficient sequence modeling~\cite{gu2023mamba,liu2024vmamba,guo2024mambair}. These approaches are typically trained using pixel-wise losses such as mean squared error, and primarily focus on enhancing restoration quality to maximize fidelity.

In recent years, the field has increasingly shifted toward models that emphasize perceptual realism. Generative approaches, such as generative adversarial networks (GANs)~\cite{goodfellow2014generative,zhang2017image,ledig2017photo}, are introduced to produce more perceptually appealing outputs. More recently, diffusion models have emerged as a powerful alternative for generating realistic textures~\cite{ho2020denoising,saharia2022image,xia2023diffir,wu2023seesr,li2024distillation}. These models progressively transform random noise into high-resolution images through a denoising process conditioned on the LR input. The diffusion-based methods show strong potential in improving perception quality, which better aligns with human visual preferences. Overall, the development of SR methods reflects a dual objective: achieving high restoration quality for faithful reconstruction, and high perception quality for producing visually pleasing, realistic results, both of which are essential for practical application.

Collaborating with the 10th edition of the NTIRE workshop at CVPR 2025, we organize the NTIRE 2025 Challenge on example-based single image super-resolution ($\times$4). This challenge focuses on recovering high-resolution details from a single low-resolution input, following the classical bicubic degradation setting. The goal is to advance the development of effective SR models and provide a unified benchmark for comparison.

To comprehensively evaluate performance, the challenge includes two tracks: one targeting restoration quality, which focuses on pixel-wise accuracy, and the other targeting perception quality, which emphasizes the perceptual realism of the output. This dual-track design reflects the evolving goals of image SR research and encourages solutions that balance fidelity with perceptual quality.

This challenge is one of the NTIRE 2025~\footnote{\url{https://www.cvlai.net/ntire/2025/}} Workshop associated challenges on: ambient lighting normalization~\cite{ntire2025ambient}, reflection removal in the wild~\cite{ntire2025reflection}, shadow removal~\cite{ntire2025shadow}, event-based image deblurring~\cite{ntire2025event}, image denoising~\cite{ntire2025denoising}, XGC quality assessment~\cite{ntire2025xgc}, UGC video enhancement~\cite{ntire2025ugc}, night photography rendering~\cite{ntire2025night}, image super-resolution (x4)~\cite{ntire2025srx4}, real-world face restoration~\cite{ntire2025face}, efficient super-resolution~\cite{ntire2025esr}, HR depth estimation~\cite{ntire2025hrdepth}, efficient burst HDR and restoration~\cite{ntire2025ebhdr}, cross-domain few-shot object detection~\cite{ntire2025cross}, short-form UGC video quality assessment and enhancement~\cite{ntire2025shortugc,ntire2025shortugc_data}, text to image generation model quality assessment~\cite{ntire2025text}, day and night raindrop removal for dual-focused images~\cite{ntire2025day}, video quality assessment for video conferencing~\cite{ntire2025vqe}, low light image enhancement~\cite{ntire2025lowlight}, light field super-resolution~\cite{ntire2025lightfield}, restore any image model (RAIM) in the wild~\cite{ntire2025raim}, raw restoration and super-resolution~\cite{ntire2025raw} and raw reconstruction from RGB on smartphones~\cite{ntire2025rawrgb}.

% \vspace{-3.mm}
\section{NTIRE 2025 Image Super-Resolution ($\times$4)}
% \vspace{-1.mm}
The NTIRE 2025 Image Super-Resolution ($\times$4) Challenge, which is one of the associated challenges of NTIRE 2025, has two primary objectives. Firstly, it intends to offer a thorough overview of the latest advancements and emerging tendencies within the field of image super-resolution (SR). Secondly, it functions as a platform that enables both academic researchers and industrial professionals to converge and investigate possible collaborative opportunities. The following part delves into the specific aspects.  

\vspace{-2.mm}
\subsection{Dataset}
\vspace{-1mm}
The challenge provides three official datasets, including DIV2K~\cite{timofte2017ntire}, Flickr2K~\cite{lim2017enhanced}, and LSDIR~\cite{li2023lsdir}. Additionally, supplementary data are allowed to be used. For the challenge, the LR-HR image pairs are constructed with the HQ images via bicubic interpolation with a $\times$4 downsampling.

\noindent\textbf{DIV2K}
The DIV2K dataset contains 1,000 2K images, which are divided into three parts: 800/100/100 for training, validation, and testing, respectively. To ensure fairness, participants do not have access to the high-resolution (HR) images from the DIV2K validation set except during the testing phase. Furthermore, the HR test images remain hidden throughout the entire challenge.

\noindent\textbf{Flickr2K}
The Flickr2K dataset has 2,650 2K images, covering a wide range of quality levels and contents. For this challenge, all images in the Flickr2K dataset are available.

\noindent\textbf{LSDIR}
The LSDIR dataset comprises 86,991 images from the Flickr platform. It is divided into three parts: 84,991/1000/1000 for training, validation, and testing.

\subsection{Track and Competition}
\label{sec:track}
This year, the competition features two tracks: the Restoration track and the Perceptual track.

\noindent\textbf{Restoration Track.} Consistent with last year’s challenge~\cite{ntire2024srx4}, all teams are ranked based on the PSNR of their enhanced HR images compared to the GT HR images from the DIV2K testing dataset. 

\noindent\textbf{Perceptual Track.} This year, a perceptual track has been introduced. For this track, seven widely used IQA metrics are collected to evaluate the restored results thoroughly. These metrics include {LPIPS, DISTS, CLIP-IQA, MANIQA, MUSIQ, and NIQE.} The teams are ultimately ranked based on the overall perceptual score:
\begin{equation}
\begin{aligned}
        \text{Score} = \left(1 - \text{LPIPS}\right) + \left(1 - \text{DISTS}\right) + \text{CLIP-IQA} \\
        + \text{MANIQA} + \frac{\text{MUSIQ}}{100} + \max\left(0, \frac{10 - \text{NIQE}}{10}\right). 
\end{aligned}
\end{equation}

\noindent\textbf{Challenge Phases.}
\textit{(1) Development and Validation Phase:} During this phase, participants have access to two datasets: (a) 800 pairs of LR/HR training images and 100 LR validation images from the DIV2K dataset, and (b) 84,991 LR/HR training image pairs and 1,000 LR validation images from the LSDIR dataset. Participants are also permitted to use additional data for training purposes. The restored HR images generated by their models can be submitted to the Codalab server for evaluation based on eight performance metrics, with immediate feedback provided.

\textit{(2) Testing Phase:} In the final phase, participants receive 100 LR test images, but the corresponding HR ground truth images are not provided. They are required to upload their SR outputs to the Codalab server and submit their code and a detailed report to the organizers via email. After the challenge ends, the organizers will validate the code and send the results to the participants.

\noindent\textbf{Evaluation Protocol.} The evaluation process involves eight standard metrics: PSNR, SSIM, LPIPS, DISTS, NIQE, ManIQA, MUSIQ, and CLIP-IQA. For the evaluation, a 4-pixel border around each image is excluded, and the calculations are performed on the Y channel of the YCbCr color space. The results are primarily based on the submissions made to the Codalab server. Code submitted by participants is used for reproduction and verification, with small discrepancies in precision being considered acceptable. A script for calculating these metrics can be found at \url{https://github.com/zhengchen1999/NTIRE2025_ImageSR_x4}, and the repository also includes the source code and pre-trained models.

\begin{table*}[t]
    \centering
    \setlength{\tabcolsep}{1.3mm}
    \begin{adjustbox}{width=\linewidth}
    % \begin{tabularx}{\linewidth}{l|rr|ccccccccx}
    \begin{tabular}{l|cc|ccccccccc}
        \toprule
        \multirow{2}{*}{Team Name} & {Rank} & {Rank} & {PSNR} & \multirow{2}{*}{SSIM } & \multirow{2}{*}{LPIPS } & \multirow{2}{*}{DISTS } & \multirow{2}{*}{NIQE } & \multirow{2}{*}{ManIQA } & \multirow{2}{*}{MUSIQ } & \multirow{2}{*}{CLIP-IQA } & {Perceptual Score} \\
        & \textbf{Track 1} & \textbf{Track 2} & \textbf{Track 1} & & & & & & & & \textbf{Track 2}  \\
        \midrule
        SamsungAICamera         &  1 &  3 & 33.46 & 0.9124 & 0.1681 & 0.0929 & 4.4089 & 0.3566 & 62.1058 & 0.4934 & 3.7692 \\
        SNUCV         & 24 &  1 & 22.53 & 0.6326 & 0.2113 & 0.1082 & 2.9635 & 0.4939 & 71.4919 & 0.7543 & 4.3472 \\
        BBox          &  2 &  4 & 31.97 & 0.8793 & 0.2082 & 0.1140 & 5.0643 & 0.3656 & 61.2975 & 0.5206 & 3.6706 \\
        MicroSR       & 22 &  2 & 26.34 & 0.7594 & 0.2340 & 0.1261 & 3.7380 & 0.3552 & 64.2008 & 0.6317 & 3.8950 \\
        XiaomiMM      &  3 &  5 & 31.93 & 0.8775 & 0.2144 & 0.1186 & 5.2336 & 0.3684 & 60.8244 & 0.5152 & 3.6355 \\
        NJU\_MCG      &  4 & 11 & 31.19 & 0.8661 & 0.2255 & 0.1233 & 5.3914 & 0.3649 & 59.9235 & 0.4984 & 3.5747 \\
        X-L           &  5 & 13 & 31.15 & 0.8653 & 0.2276 & 0.1227 & 5.3885 & 0.3597 & 59.5546 & 0.4990 & 3.5652 \\
        Endeavour     &  6 & 12 & 31.15 & 0.8653 & 0.2269 & 0.1229 & 5.3787 & 0.3610 & 59.6981 & 0.4994 & 3.5697 \\
        CidautAi      & 20 &  6 & 30.48 & 0.8564 & 0.2029 & 0.0894 & 4.7080 & 0.3384 & 59.3074 & 0.4472 & 3.6157 \\
        KLETech-CEVI  &  7 &  7 & 31.13 & 0.8649 & 0.2264 & 0.1217 & 5.3118 & 0.3661 & 60.0893 & 0.5087 & 3.5964 \\
        JNU620        &  8 & 10 & 31.12 & 0.8647 & 0.2271 & 0.1212 & 5.3608 & 0.3587 & 59.6759 & 0.5048 & 3.5758 \\
        ACVLAB        & 14 &  8 & 30.82 & 0.8635 & 0.2302 & 0.1210 & 5.2777 & 0.3642 & 59.9242 & 0.5071 & 3.5916 \\
        CV\_SVNIT     &  9 &  9 & 31.11 & 0.8647 & 0.2259 & 0.1224 & 5.3500 & 0.3619 & 59.8108 & 0.5010 & 3.5778 \\
        HyperPix      & 10 & 14 & 31.03 & 0.8633 & 0.2286 & 0.1238 & 5.3826 & 0.3593 & 59.5252 & 0.4982 & 3.5621 \\
        BVIVSR        & 11 & 15 & 30.99 & 0.8630 & 0.2288 & 0.1234 & 5.4281 & 0.3602 & 59.6739 & 0.4969 & 3.5588 \\
        AdaDAT        & 12 & 17 & 30.91 & 0.8605 & 0.2329 & 0.1256 & 5.4721 & 0.3584 & 59.1563 & 0.4968 & 3.5410 \\
        Junyi         & 13 & 19 & 30.91 & 0.8605 & 0.2364 & 0.1278 & 5.5369 & 0.3542 & 58.7826 & 0.4926 & 3.5167 \\
        ML\_SVNIT     & 15 & 20 & 30.82 & 0.8589 & 0.2357 & 0.1268 & 5.4299 & 0.3497 & 58.4427 & 0.4831 & 3.5117 \\
        SAK\_DCU      & 16 & 16 & 30.80 & 0.8595 & 0.2328 & 0.1268 & 5.3981 & 0.3551 & 59.2546 & 0.4937 & 3.5419 \\
        VAI-GM        & 17 & 18 & 30.76 & 0.8579 & 0.2366 & 0.1279 & 5.4573 & 0.3496 & 58.8017 & 0.4919 & 3.5193 \\
        Quantum Res   & 18 & 22 & 30.52 & 0.8523 & 0.2482 & 0.1330 & 5.6358 & 0.3386 & 56.9043 & 0.4754 & 3.4381 \\
        PSU           & 19 & 21 & 30.50 & 0.8528 & 0.2476 & 0.1296 & 5.5450 & 0.3404 & 57.6043 & 0.4801 & 3.4649 \\
        IVPLAB-sbu    & 21 & 23 & 26.74 & 0.8490 & 0.4512 & 0.1992 & 5.4675 & 0.3384 & 56.2507 & 0.4890 & 3.1927 \\
        MCMIR         & 23 & 24 & 24.53 & 0.7220 & 0.2624 & 0.1460 & 6.1282 & 0.2773 & 42.8159 & 0.3456 & 3.0298 \\
        \midrule
        Aimanga       & N/A & N/A & 22.44 & 0.6553 & 0.2965 & 0.1396 & 2.8260 & 0.4752 & 68.8273 & 0.7333 & 4.1781 \\
        IPCV\_Team    & N/A & N/A & 31.01 & 0.8643 & 0.2255 & 0.1228 & 5.3553 & 0.3664 & 60.2005 & 0.5033 & 3.5878 \\
        \bottomrule
    % \end{tabularx}
    \end{tabular}
    \end{adjustbox}
    \caption{\textbf{Results of NTIRE 2025 Image Super-Resolution Challenge.} PSNR and the other seven perceptual metric scores are measured on the DIV2K testing (100) dataset. For the restoration track, the team rankings are based primarily on PSNR. As for the perceptual track, the team ranking is based on the perceptual score, a weighted average of seven perceptual metrics. 
    The ranking mainly depends on the higher ranking among the two tracks, and secondly on the average value of the rankings in the two tracks.
    Team order in Sec.~\ref{sec:teams} follows the same sequence as this table. Teams (\ie, Aimanga and IPCV\_Team) that submitted late or failed to submit their code are not ranked.}
    \label{tab:main_results}
    % \vspace{-4.mm}
\end{table*}

% \vspace{-2mm}
\section{Challenge Results}
% \vspace{-1mm}
The challenge includes two sub-tracks: Track 1 (restoration quality) and Track 2 (perception quality). The results and rankings of all participating teams are provided in Tab.~\ref{tab:main_results}.

\noindent\textbf{Track 1 (Restoration Quality).} The SamsungAICamera team achieves the top performance (33.46 dB). Notably, two teams surpass last year’s best PSNR score (31.94 dB), and ten teams obtain results above 31.00 dB, highlighting a clear improvement in reconstruction accuracy.

\noindent\textbf{Track 2 (Perception Quality).} The SNUCV team ranks first with the highest perceptual score (4.3472). Two teams achieve a score above 4.0, and seven teams exceed 3.6, indicating advancements in the perceptual quality.

Further details on the evaluation protocol of the two tracks are provided in Sec.~\ref{sec:track}. The team order in Sec.~\ref{sec:teams} follows the same order as presented in Tab.~\ref{tab:main_results}, with the top 5 teams and their method details highlighted here. Due to space limitations, the remaining teams are listed in Sec.~A of the supplementary materials. Team member information can also be found in Sec.~B of the supplementary materials. 

\subsection{Architectures and main ideas}
Throughout the challenge, various innovative techniques were introduced to boost the SR performance. Here, the team members summarize some of the principal concepts.

\begin{enumerate}
    \item \textbf{Transformer-based architectures remain a mainstream approach.} Transformer-based methods, such as HAT~\cite{chen2023activating}, SwinIR~\cite{liang2021swinir}, and DAT~\cite{chen2023dual}, continue to deliver strong reconstruction results by capturing long-range dependencies. Many teams utilized pre-trained Transformer models and fine-tuned them with hybrid attention and self-attention mechanisms. The BBox team, for example, used Transformer models to capture both local and global information, improving reconstruction quality through model ensembles.  

    \vspace{2.mm}
    \item \textbf{Integrating Mamba architectures for improved global context modeling.} Recognizing the importance of capturing extensive contextual information, some participants employed the Mamba architecture to better model long-range spatial dependencies and global context, leading to significant performance improvements in super-resolution tasks. The XiaomiMM team, for example, integrated Mamba into the HAT model, adapting branches based on image statistics and combining multiple Mamba-enhanced networks.

    \vspace{2.mm}
    \item \textbf{Dynamic fusion of multiple network architectures boosts performance.} Several teams utilized dynamic model fusion strategies, combining outputs from Transformer-based models and CNNs. By leveraging the strengths of both architectures, these methods captured local textures and global structural information, significantly improving reconstruction quality. The winning team, SamsungAICamera, for instance, combined a pre-trained HAT Transformer with the CNN-based NAFNet, achieving top performance in the restoration track.

    \vspace{2.mm}
    \item \textbf{Incorporating frequency-domain loss functions for enhanced detail reconstruction.} To preserve fine details and textures, participants employed frequency-domain loss functions, such as Stationary Wavelet Transform (SWT) loss, to improve high-frequency detail recovery and reduce artifacts. The KLETech-CEVI team, for example, used wavelet-based loss in HAT framework, applying Symlet filters to maintain texture sharpness and prevent chroma distortion.

    \vspace{2.mm}
    \item \textbf{Advanced multi-stage training strategies to progressively refine model performance.} Participants adopted multi-stage training methods, gradually increasing patch sizes during training, which improved reconstruction quality and model stability. For example, the BBox team used a three-stage training strategy, adjusting patch sizes and loss functions across stages, followed by CLIP-based semantic filtering and fine-tuning, ultimately enhancing both fidelity and perceptual quality.

\end{enumerate}

% \vspace{-1mm}
\subsection{Participants}
% \vspace{-0.5mm}
This year, the image SR challenge attracted 286 registered participants, with 25 teams submitting valid entries. In comparison to last year's challenge~\cite{ntire2024srx4}, there was an increase in the number of valid submissions (from 20 to 24) and a significant improvement in the results. These submissions have set a new benchmark for the state-of-the-art in image super-resolution ($\times$4).

% \vspace{-1mm}
\subsection{Fairness}
% \vspace{-0.5mm}
A set of rules has been established to ensure the fairness of the competition. \textbf{(1)} The use of DIV2K test HR images for training is strictly prohibited, although the DIV2K test LR images are available. \textbf{(2)} Participants are allowed to train using additional datasets, such as the Flickr2K and LSDIR datasets. \textbf{(3)} The use of data augmentation techniques during both training and testing is considered a fair practice.

\subsection{Conclusions}
The insights gained from analyzing the results of the NTIRE 2025 image super-resolution (SR) challenge are summarized as follows:
\begin{enumerate}
    \item The integration of Transformer and CNN hybrid architectures has demonstrated exceptional performance in modeling global context and reconstructing local details, achieving a balanced approach.
    
    \item Mamba-based state space models have been widely adopted, showcasing their scalability in computational efficiency and modeling capabilities, offering a robust alternative to traditional self-attention mechanisms.
    
    \item Advanced training strategies, including multi-stage pipelines, progressive patch training, and CLIP-based semantic filtering, have boosted model generalization and robustness. Frequency-domain losses further enhanced high-frequency texture restoration.
    
    \item The innovative use of generative priors, particularly pre-trained diffusion models combined with CLIP-based reference-free perceptual losses, has achieved superior perceptual quality with minimal training.
    
\end{enumerate}

\section{Challenge Methods and Teams}
\label{sec:teams}

% Team 17	
\subsection{SamsungAICamera}
\begin{figure}
  \centering
  \includegraphics[width=0.55\linewidth]{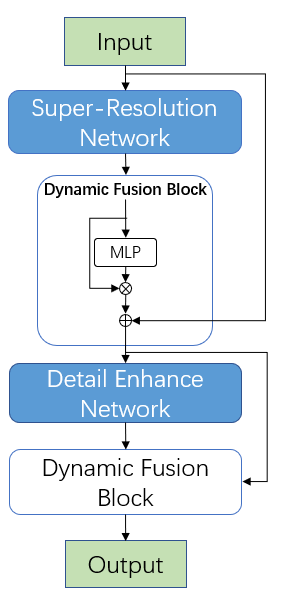}
  \vspace{-5mm}
  \caption{\textbf{Team SamsungAICamera}}
  \label{fig:Team17-fig1}
  \vspace{-4mm}
\end{figure}
\noindent\textbf{Description.}
The proposed solution is shown in Fig. \ref{fig:Team17-fig1}.
Image super-resolution (SR) is a critical task in computer vision, 
aiming to reconstruct high-quality images from their low-resolution versions. 
While significant advances have been made in recent years, 
existing methods often face challenges in effectively capturing both global context and fine local details simultaneously.

\begin{figure}
  \centering
  \includegraphics[width=1.0\linewidth]{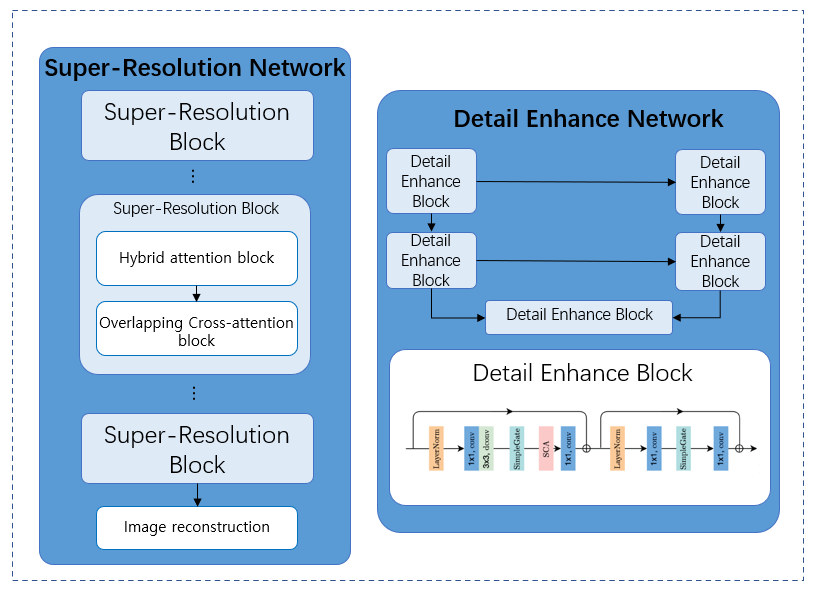}
  \vspace{-7mm}
  \caption{\textbf{Team SamsungAICamera}}
  \label{fig:Team17-SamsungAICamera-fig2}
  \vspace{-6mm}
\end{figure}

To address these limitations,  SamsungAICamera's approach combines two powerful networks: 
the transformer-based network HAT~\cite{chen2023activating} and the convolution-based network NAFnet~\cite{chen2022simple}. 
HAT try to leverage global and local information more effective by combine channel attention and overlapping cross-attention, but still struggle for 
 capturing fine-grained local features. This team design the Super-Resolution Network and Detail Enhance Network refer to these two method as shown in Fig.~\ref{fig:Team17-SamsungAICamera-fig2}.
They achieve superior performance in image super-resolution, by dynamicly fuse the feature extracted by transformer-based network and convolution-based network
This design not only addresses 
the limitations of each individual component but also opens new possibilities
 for hybrid architectures in image super-resolution tasks.

\noindent\textbf{Implementation Details.}
{\textit{Datasets.}}The team used three datasets in total: the DIV2K dataset, the LSDIR dataset, and a self-collected custom dataset consisting of 2 million images. The specific ways in which the team utilized these training sets across different training phases will be detailed in the training details section.

\begin{figure*}[t]
    \centering
    \includegraphics[width=\textwidth]{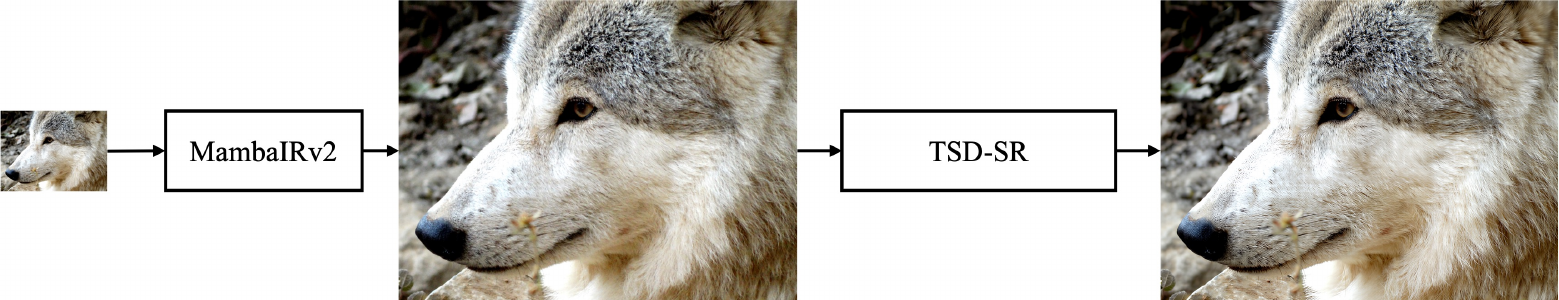}
    \caption{\textbf{Team SNUCV}}
    \label{fig:team04}
    \vspace{-6mm}
\end{figure*}

{\textit{Trainning strategy.}} The model training consists of two stages. In the first stage, The team pre-train the entire network using the custom dataset of 2 million images and LSDIR datasets, 
with an initial learning rate of $1e^{-4}$ and a training time of approximately 360 hours. 
In the second stage, they fine-tune the network module using the DIV2K datasets, with an initial learning rate of $1e^{-5}$ and a training duration of about 120 hours, 
which enhanced the model's ability to restore details.

They used data augmentation methods such as channel shuffle and mixing and perform progressive learning
where the network is trained on different image patch sizes
gradually enlarged from 320 to 448 and 768. As the patch
size increases, the performance can gradually improve.
As for training loss, they train the model by alternately iterating L1 loss, L2 loss, and Stationary Wavelet Transform(SWT) loss~\cite{korkmaz2024training}.
They found that adding SWT loss during training helps the model escape from local optima. Their model was trained on an A100 80G GPU.

\subsection{SNUCV}

\noindent\textbf{Description.}
As illustrated in Fig.~\ref{fig:team04}, the SNUCV team first employs a classic super-resolution (SR) model to perform $4\times$ upsampling of the input image. Subsequently, a one-step diffusion model is applied to further enhance the super-resolution process. For the $4\times$ upsampling model, the team adopts MambaIRv2~\citep{guo2024mambairv2}, while the diffusion model utilizes the TSD-SR~\citep{dong2024tsd} architecture. To fully leverage the generative prior of the diffusion model, they refrain from additional training and directly use its pretrained weights, focusing instead on fine-tuning the upsampler.

Let the diffusion model be \(M\), the upsampler be \(U\), the input image be \(I\), and the ground-truth image be \(I_{gt}\). To ensure that both the output \(M(U(I))\) and the intermediate output \(U(I)\) accurately resemble the ground-truth \(I_{gt}\), the team defines a loss function composed of the LPIPS loss~\citep{zhang2018unreasonable}, denoted \(\mathcal{L}_{\text{LPIPS}}\), and the L1 loss, denoted \(\mathcal{L}_{1}\):

\begin{align}
\mathcal{L}_{\text{accuracy}} &= \mathcal{L}_{1}(U(I), I_{gt}) \nonumber \\
&\quad + \alpha \cdot \mathcal{L}_{\text{LPIPS}}(M(U(I)), I_{gt}),
\end{align}

where \(\mathcal{L}_{\text{LPIPS}}\) captures perceptual similarity between the final output and the ground-truth, and \(\mathcal{L}_{1}\) enforces pixel-wise accuracy.

To further enhance perceptual quality, the team introduces a no-reference CLIP-based image quality assessment (IQA) loss~\citep{radford2021learning,wang2022exploring}, denoted \(\mathcal{L}_{\text{CLIP}}\). This loss utilizes the CLIP model's text encoder \(\text{CLIP}_{\text{text}}\) and image encoder \(\text{CLIP}_{\text{image}}\). Two text prompts are defined: ``Good photo'' and ``Bad photo'', and their embeddings are computed and normalized as:

\begin{align}
T_{\text{pos}} &= \text{norm}(\text{CLIP}_{\text{text}}(\text{``Good photo"})) \\
T_{\text{neg}} &= \text{norm}(\text{CLIP}_{\text{text}}(\text{``Bad photo"}))
\end{align}

The generated image's feature representation is:
\begin{equation}
F_{\text{pred}} = \text{norm}(\text{CLIP}_{\text{image}}(M(U(I))))
\end{equation}

Then, the similarities between the image features and the text embeddings are computed:
\begin{align}
S_{\text{pos}} &= F_{\text{pred}} \cdot T_{\text{pos}}^{\top} \\
S_{\text{neg}} &= F_{\text{pred}} \cdot T_{\text{neg}}^{\top}
\end{align}

The CLIP loss is defined to increase similarity with ``Good photo'' and reduce similarity with ``Bad photo'':

\begin{equation}
\mathcal{L}_{\text{CLIP}} = 1 - S_{\text{pos}} + S_{\text{neg}}
\end{equation}

The total training loss is formulated as:
\begin{equation}
\mathcal{L} = \mathcal{L}_{\text{accuracy}} + \beta \cdot \mathcal{L}_{\text{CLIP}}
\end{equation}

where they set \(\alpha = 0.1\) and \(\beta = 0.5\).

\noindent\textbf{Implementation Details.}
When training the upsampler MambaIRv2, the team utilizes the AdamW~\citep{loshchilov2017decoupled} optimizer with hyperparameters \(\beta_1 = 0.9\), \(\beta_2 = 0.999\), and a learning rate of \(1 \times 10^{-5}\). The ground-truth images are cropped into patches of size \(256 \times 256\). Training is conducted over 100,000 iterations with a batch size of 4. The training dataset is DIV2K~\citep{timofte2017ntire} and LSDIR~\citep{li2023lsdir}.

\begin{figure*}[t]
  \centering
  \includegraphics[width=0.9\textwidth]{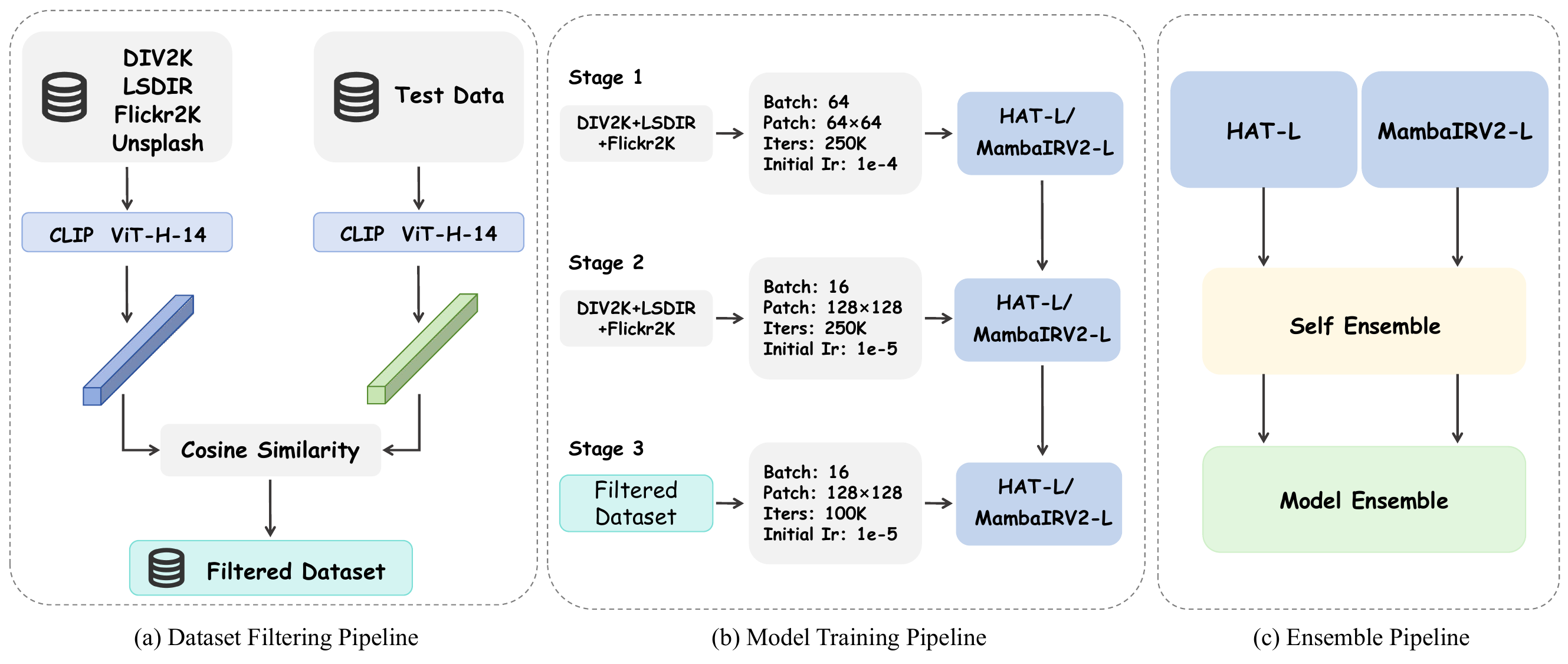}
  \caption{\textbf{Team BBox.}}
  \label{fig::pipeline}
  \vspace{-6mm}
\end{figure*}

% Team 15	
\subsection{BBox}
\noindent\textbf{Description.}
The solution proposed by the BBox is illustrated in Fig.~\ref{fig::pipeline}. Transformer-based models have consistently demonstrated remarkable performance in the field of super-resolution, as exemplified by methods such as HAT~\cite{chen2023activating}, DAT~\cite{chen2023dual}, and SwinIR~\cite{liang2021swinir}. 
Recently, numerous studies have shown that Mamba architectures can also achieve impressive results in this domain, as evidenced by models such as MambaIR~\cite{guo2024mambair}, MambaIRv2~\cite{guo2024mambairv2}, and S$^3$Mamba~\cite{Xia2024s3mamba}. 
Notably, prior research indicates that Transformer-based models excel at modeling sequential relationships, while Mamba-based approaches are particularly adept at capturing long-range contextual information~\cite{ntire2024srx4}. 
Both capabilities are crucial for pixel-dense super-resolution tasks, which require models to simultaneously capture spatial relationships between individual pixels and model their long-range contextual dependencies. 
Inspired by these complementary strengths, the team members adopt HAT-L and MambaIRv2-L as the foundation models, training them independently and employing a model ensemble strategy to effectively leverage their advantages, thereby achieving optimal performance. 

\noindent\textbf{Implementation Details.} 
The training dataset comprises DIV2K, LSDIR, Flickr2K, and selected Unsplash~\cite{UnsplashData} datasets. To augment the training data, the team members implement random flip and rotation strategies. For the HAT model, the team members initialize the network using pre-trained weights from the HAT-L model, which was previously trained on the ImageNet dataset. In contrast, the team members train the MambaIRv2-L model from scratch.

\begin{figure}[t]
    \centering
    \includegraphics[width=\linewidth]{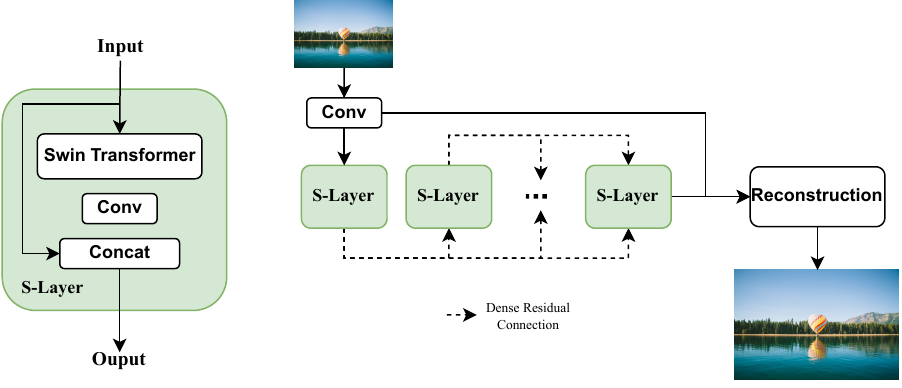}
    \caption{\textbf{Team MicroSR}}
    \label{fig:team07}
    \vspace{-6mm}
\end{figure}

\begin{figure*}[t]
\centering
\includegraphics[width=1.0\textwidth]{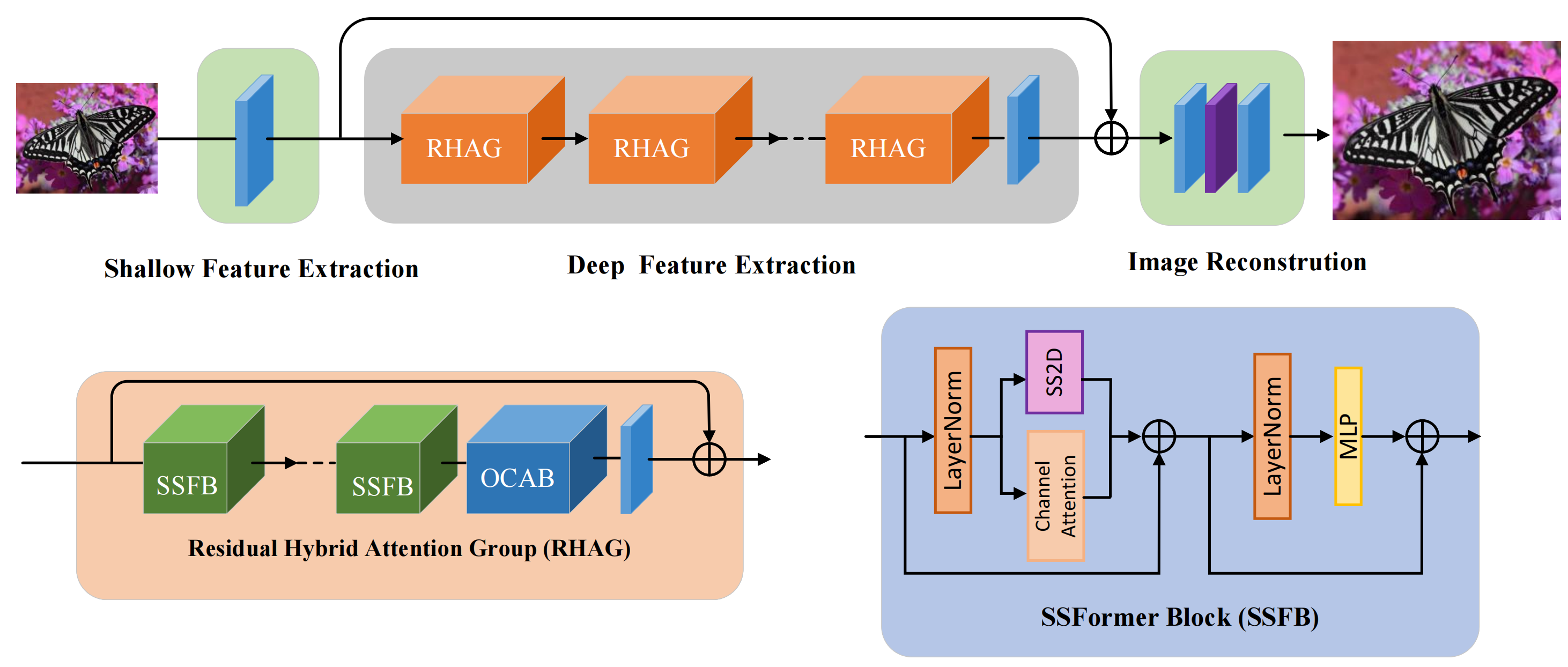}
\caption{\textbf{Team XiaomiMM}}
\vspace{-6mm}
\label{fig:team18}
\end{figure*}

Specifically, the training follows a three-stage progressive strategy with a multi-scale loss that computes losses at x2, x3, and x4 resolutions to enhance performance across different scales. (1) In the first stage, the team members train on the DIV2K, LSDIR, and Flickr2K datasets. The team members set the patch size to 64 and batch size to 64, conducting 250K iterations with an initial learning rate of $1\times10^{-4}$. During this stage, the team members minimize the L1 loss using the Adam optimizer~\cite{kingma2014adam}.
(2) In the second stage, while maintaining the same training datasets, the team members increase the patch size to 128 and reduce the batch size to 16. The team members also switch from L1 loss to MSE loss. This stage comprises 250K iterations with a reduced initial learning rate of $1\times10^{-5}$.
(3) In the final stage, to achieve superior model performance, the team members implement semantic selection using CLIP~\cite{radford2021learning} features. the team members filter out 2,000 similar images from DIV2K, LSDIR, Flickr2K, and Unsplash datasets for further fine-tuning. During this stage, the learning rate is set to $1\times10^{-5}$, and the model is trained for an additional 100K iterations.
The multi-step learning rate decay method is applied across all three training phases.
For both MambaIRv2-L and HAT-L models, the team members repeat the aforementioned three training stages five times to ensure optimal convergence.

In the inference stage, the team members employ a self-ensemble strategy to enhance the performance of both HAT-L and MambaIRv2-L models. For MambaIRv2-L specifically, the team members utilize multiple sliding windows of varying sizes during inference and integrate these results. Finally, the team members implement a weighted fusion method~\cite{2023ntire} to generate outputs from the adaptive combination of HAT-L and MambaIRv2-L models.

% Team 7	
\vspace{-2mm}
\subsection{MicroSR}
\vspace{-1mm}

\noindent\textbf{Description.}
The MicroSR team proposes a super-resolution method based on the architectural principles of DRCT~\cite{hsu2024drct} and SwinFIR~\cite{zhang2022swinfir}, while also building on the foundation established by SwinIR~\cite{liang2021swinir}. A novel S-layer is introduced with residual connections to improve the capture of global information more efficiently. As illustrated in Fig.~\ref{fig:team07}, the architecture comprises a shallow feature extraction layer, 12 S-layers constructed with dense residual connections, and a final reconstruction layer.

\noindent\textbf{Implementation Details.}
The proposed solution adopts a multi-stage greedy training policy. In the first stage, the model is trained with MSE loss for 10 epochs. In the second stage, an adversarial loss is introduced to fine-tune the model for better perceptual quality. During training, the learning rate decays progressively by a factor of 0.1 after each stage. Within each stage, the learning rate remains fixed, and the best-performing checkpoint is selected and used to initialize the next stage.

To further enhance performance, the team integrates a neural degradation algorithm~\cite{luo2024and} for data augmentation. The training data includes the DIV2K dataset~\cite{timofte2017ntire}, supplemented with real-world paired images~\cite{guo2022data}, resulting in a more diverse dataset and improved robustness to unseen degradations. The team re-trains DRCT using the official training code on the augmented dataset. Training is initialized with the official pretrained model Real-DRCT-GAN\_Finetuned. The final model contains 27.58M parameters, with 11.078G FLOPs. The average inference time per image is reported as 4.41 seconds.

% Team 18	
\subsection{XiaomiMM}

\noindent\textbf{Description.}
Based on the HAT model~\cite{chen2023activating}, the team members propose two variants to improve image super-resolution performance.

The first variant introduces a multi-branch structure, designed according to the statistical feature information (brightness, contrast, sharpness, noise, saturation) of the input image. Each of these five statistical features is divided into two categories, resulting in a ten-branch network structure. By utilizing fixed statistical information instead of allowing the network to learn these features autonomously, the team members introduce additional prior information and improve the model's generalization ability.

The second variant integrates the Mamba structure~\cite{liu2024vmamba} into the HAT model, thereby fully leveraging the advantageous characteristics of Mamba. The detailed network structure is illustrated in Fig.~\ref{fig:team18}.

Finally, the team members trained and fused the original HAT model and these two variants, achieving the best results in the experiments.

\noindent\textbf{Implementation Details.} 
The dataset utilized for training comprises DIV2K and LSDIR. During each training batch, 64 HR RGB patches measuring $256 \times 256$ are cropped and subjected to random flipping and rotation. The learning rate is initialized at $5\times 10^{-4}$ and is halved every $2\times10^5$ iterations. The network undergoes training for a total of $10^6$ iterations, minimizing the L1 loss function using the Adam optimizer~\cite{kingma2014adam}. The team members repeated the aforementioned training settings four times after loading the trained weights. Subsequently, fine-tuning is executed using the L1 and L2 loss functions, with an initial learning rate of $1\times10^{-5}$ for $5\times10^5$ iterations and an HR patch size of 512. The team members conducted fine-tuning on four models utilizing both L1 and L2 losses and employed batch sizes of 64 and 128. Finally, the team members integrated these five models to obtain the ultimate model and also used the K-means Based Fusion Strategy~\cite{ntire2024srx4} to achieve better results.

\vspace{-2.mm}
\section{Methods of the Remaining Teams}
\vspace{-2.mm}
All the teams presented innovative ideas and thorough experiments for the competition. However, due to space limitations, a more in-depth discussion can be found in Sec.~A of the supplementary materials, which contains detailed descriptions of the methods and implementation details for the remaining teams that participated in the challenge. While these teams have not been discussed in the main report, their approaches are still highlighted, offering insight into their unique strategies and technical implementations.

\vspace{-2.mm}
\section*{Acknowledgements}
\vspace{-2.mm}
This work is supported by the Shanghai Municipal Science and Technology Major Project (2021SHZDZX0102) and the Fundamental Research Funds for the Central Universities.
This work was partially supported by the Humboldt Foundation. We thank the NTIRE 2025 sponsors: ByteDance, Meituan, Kuaishou, and University of Wurzburg (Computer Vision Lab).

%%%%%%%%% REFERENCES
{\small
\bibliographystyle{ieeenat_fullname}
\bibliography{main}
}

\end{document}